\documentclass{article}
\usepackage{diagbox} 

\PassOptionsToPackage{table,xcdraw}{xcolor}

\usepackage{algorithm}
\usepackage{algorithmic}
\usepackage{chngcntr}
\counterwithout{algorithm}{section}

\usepackage{amsmath}
\usepackage{amssymb}
\usepackage{mathtools}

\usepackage[final]{neurips_2024}

\usepackage[utf8]{inputenc} 
\usepackage[T1]{fontenc}    
\usepackage{hyperref}       
\usepackage{url}            
\usepackage{booktabs}       
\usepackage{amsfonts}       
\usepackage{nicefrac}       
\usepackage{microtype}      
\usepackage{xcolor}         

\title{J6: Jacobian-Driven Role Attribution for Multi-Objective Prompt Optimization in LLMs}

\author{%
  Yao Wu\\ 
  School of Engineering \\ 
  Westlake University \\ 
  Hangzhou, China \\ 
  \texttt{wuyao@westlake.edu.cn} \\ 
}

\begin{document}

\maketitle

\begin{abstract}
  In large language model (LLM) adaptation, balancing multiple optimization objectives—such as improving factuality (\emph{heat}) and increasing confidence (via low entropy)—poses a fundamental challenge, especially when prompt parameters (e.g., hidden-layer insertions $h$ and embedding modifications $w$) interact in non-trivial ways. Existing multi-objective optimization strategies often rely on scalar gradient aggregation, ignoring the deeper geometric structure between objectives and parameters. We propose \textbf{J6}, a structured Jacobian-based method that decomposes the gradient interaction matrix into six interpretable components: $[ \|J_{11}\|^2,\ \langle J_{11}, J_{22} \rangle,\ \|J_{12}\|^2,\ \|J_{21}\|^2,\ \|J_{22}\|^2,\ \langle J_{21}, J_{12} \rangle]$. This decomposition enables both \emph{hard decision-making} (e.g., choosing the dominant update direction via $\arg\max$) and \emph{soft strategies} (e.g., attention-style weighting via softmax over J6), forming a dynamic update framework that adapts to local conflict and synergy. Moreover, the interpretable structure of J6 provides insight into parameter attribution, task interference, and geometry-aligned adaptation. Our work introduces a principled and extensible mechanism for conflict-aware prompt optimization, and opens a new avenue for incorporating structured Jacobian reasoning into multi-objective neural tuning.
\end{abstract}

\section{Introduction}

Large Language Models (LLMs) are increasingly deployed in real-world scenarios that demand not only high accuracy (e.g., factual correctness), but also strong reliability (e.g., output certainty). This dual requirement gives rise to a growing need for \textit{multi-objective prompt optimization}, where practitioners aim to improve both \textbf{fidelity} and \textbf{certainty} simultaneously—two goals that are often at odds during model inference~\cite{nema2025modp,wang2024rega}.

Existing prompt tuning approaches typically optimize a single scalar loss or apply fixed-weight multi-task objectives. However, these approaches treat all prompt parameters homogeneously and fail to model the underlying structure between different parameter types and their contributions to different objectives~\cite{dun2023mops}. Specifically, they overlook the fact that different perturbation types—such as hidden-layer perturbations ($h$) and embedding perturbations ($w$)—may affect each objective differently, and often nonlinearly, due to the multiplicative nature of the logit computation: $(H + h)(W + w)^T$.

A natural yet flawed solution is to statically assign $h$ to optimize \texttt{fidelity} (i.e., cross-entropy minimization) and $w$ to optimize \texttt{certainty} (i.e., entropy minimization). However, such a rigid role assignment is overly simplistic and brittle in practice. Both $h$ and $w$ may influence both objectives in entangled and dynamic ways~\cite{wang2024rega}. A hidden-layer perturbation that improves answer accuracy may simultaneously disrupt confidence, and vice versa. In such settings, pre-assigning roles leads to suboptimal updates and gradient conflicts.

We argue for a fundamentally different perspective: instead of statically binding $h$ and $w$ to fixed tasks, we enable \textit{dynamic and competitive role attribution}, where each parameter group autonomously decides—at each optimization step—whether to respond to one objective, both, or neither. This allows parameters to behave as adaptive agents that \textit{sniff} task relevance based on local gradient signals~\cite{quinton2025jacobian}.

To implement this, we introduce \textbf{J6}, a Jacobian-based attribution framework that reveals the fine-grained interaction structure between objectives and parameter groups. At each step, we compute a local $2 \times 2$ Jacobian matrix $J$ over two objectives (\texttt{heat} and \texttt{confidence}) and two parameter groups ($h$, $w$), and decompose $J$ into six interpretable components:
\[
    \text{J6} =
    \left[
    \underbrace{\|J_{11}\|^2}_{\text{h} \rightarrow \text{Heat}}, \quad
    \underbrace{\langle J_{11}, J_{22} \rangle}_{\text{Alignment}}, \quad
    \underbrace{\|J_{12}\|^2}_{\text{w} \rightarrow \text{Heat}}, \quad
    \underbrace{\|J_{21}\|^2}_{\text{h} \rightarrow \text{Conf}}, \quad
    \underbrace{\|J_{22}\|^2}_{\text{w} \rightarrow \text{Conf}}, \quad
    \underbrace{\langle J_{21}, J_{12} \rangle}_{\text{Alignment}}
    \right]
\]
Each J6 entry quantifies either a role-specific contribution (e.g., $h \rightarrow \texttt{heat}$) or a structural interaction (e.g., cross-objective alignment) within the gradient space.

Building on this decomposition, we propose two update schemes:

\begin{itemize}
    \item \textbf{Hard Strategy}: Select the maximal entry in the J6 vector to determine which parameter to update and which objective to prioritize. This produces interpretable, discrete role assignments at each step.
    \item \textbf{Soft Strategy}: Normalize J6 scores via softmax and use them as continuous weights to guide simultaneous updates across $h$ and $w$, allowing gradient blending without dimensional expansion.
\end{itemize}

Beyond J6, we further extend this idea to a richer gradient attribution space $\text{J}^+$, comprising 15 high-order alignment terms that account for cross-role synergy, intra-role consistency, and asymmetric dominance. This extended view is not used directly in training, but provides the theoretical foundation for our Soft Strategy. Notably, we show that a properly weighted J6 vector is sufficient to emulate the optimization behavior of $\text{J}^+$—offering high expressivity without incurring extra computation.

Empirically, we evaluate J6 on three benchmarks—\texttt{MathQA}, \texttt{GSM8K}, and \texttt{TruthfulQA}—under settings with severe objective conflicts. Our method consistently outperforms strong baselines including \texttt{Slot}, \texttt{PCGrad}, and Pareto-based methods~\cite{nema2025modp}, while offering insightful visualizations of how $h$ and $w$ adapt roles over time. These findings demonstrate that Jacobian-driven role attribution is not only effective, but also interpretable and extensible.

\vspace{0.5em}
\textbf{Contributions.} This paper makes the following key contributions:
\begin{enumerate}
    \item We introduce the problem of role attribution in multi-objective prompt tuning, and formalize it through Jacobian decomposition~\cite{quinton2025jacobian}.
    \item We propose J6, a simple yet expressive mechanism that supports both hard (argmax) and soft (weighted) update strategies.
    \item We extend the attribution space to a 15-term vector $\text{J}^+$, and show that soft-weighted J6 suffices to emulate it.
    \item We demonstrate strong empirical results on three tasks, along with interpretable analysis of parameter-objective dynamics.
\end{enumerate}

\section{Related Work}

\subsection{Multi-Objective Optimization in Deep Learning}

Balancing multiple, often conflicting, objectives has been a long-standing challenge in multi-task and multi-objective learning. A classic solution is PCGrad~\cite{yu2020gradient}, which reduces destructive interference by projecting gradients onto the normal plane of each other. ParetoMTL~\cite{lin2019pareto} explores Pareto frontiers to generate non-dominated trade-offs across objectives. Geometry-aware approaches~\cite{sener2018multi} further introduce vector projections and alignment constraints to decouple task interactions.

While effective at the global task level, these methods typically operate on coarse-grained gradient representations and overlook the heterogeneous roles that different parameter groups may play. In contrast, our approach performs \textit{local, parameter-level decomposition} via a Jacobian-based scoring mechanism, enabling fine-grained control over optimization responsibilities within the model.



\subsection{Prompt Tuning and Delta-Based Adaptation}

Prompt tuning has emerged as a lightweight alternative to full fine-tuning, allowing efficient adaptation of LLMs with minimal additional parameters~\cite{lester2021power}. Methods such as Prefix Tuning~\cite{li2021prefix} and LoRA~\cite{hu2021lora} introduce trainable vectors or low-rank matrices into intermediate activations or weight spaces. Delta Tuning~\cite{ding2022delta} generalizes this idea by enabling parameter insertions at arbitrary locations, making adaptation more modular and extensible.

\subsection*{SLOT Method}

\textbf{SLOT}~\cite{hu2025slotsamplespecificlanguagemodel} introduces instance-level adaptation at test time by injecting a perturbation into the hidden state of the LLM. Given the final hidden state $H$ and output projection matrix $W$, SLOT modifies the logits as:
\[
\text{logits}' = (H + \delta) W^\top
\]
where $\delta$ is a test-time learned perturbation, optimized via cross-entropy over the same prompt:
\[
\delta = \arg\min_{\delta} \text{CrossEntropy}((H + \delta) W^\top, y)
\]
This hidden-layer intervention tunes the model's behavior without retraining, improving sample-specific alignment.

This optimization encourages the model to better align its internal representations with prompt intent, serving as an effective one-shot alignment strategy under limited supervision. 
Despite their flexibility, these methods typically collapse multi-objective behaviors (e.g., accuracy and confidence) into a single scalar loss, making them vulnerable to hidden conflicts among competing goals. For example, a perturbation that enhances token prediction (fidelity) might inadvertently inflate entropy (uncertainty).

To overcome this limitation, we propose J6, a Jacobian-guided attribution mechanism that explicitly separates gradient responsibilities across multiple objectives. Unlike SLOT's single-loss formulation, J6 empowers parameters such as $h$ and $w$ to dynamically specialize in response to distinct objectives, Heat (accuracy) and Confidence (certainty), based on fine-grained Jacobian signals. This allows for more interpretable, role-specific optimization and mitigates gradient interference during multi-objective prompt adaptation.

\subsection*{EM-INF}

\textbf{EM-INF}~\cite{agarwal2025unreasonableeffectivenessentropyminimization} directly optimizes entropy at the logits layer:
\[
\mathcal{L}_{\text{entropy}} = -\mathrm{Entropy}(\mathrm{softmax}(\text{logits}))
\]
By targeting certainty as an independent goal, EM-INF enhances confidence in complex reasoning tasks. Yet, this can degrade factual accuracy if confidence diverges from correctness.

Our method draws from EM-INF’s focus on entropy but embeds it into a two-objective system. We use the Jacobian structure to allocate optimization pressure across parameters, allowing fidelity and certainty to co-adapt rather than compete blindly.

\subsection{Jacobian-Based Modeling and Optimization}

Jacobian matrices have long been used to analyze neural network sensitivity~\cite{sundararajan2017axiomatic}, generalization via neural tangent kernels~\cite{jacot2018neural}, and regularization~\cite{drucker1992improving}. More recently, per-layer Jacobians have been applied to investigate learning dynamics~\cite{fort2021drawing} and to improve representation alignment in multi-view learning~\cite{navon2023multi}.

However, existing works primarily use Jacobians as static diagnostics or post-hoc analytical tools. In contrast, our work treats the Jacobian as a \textit{first-class optimization signal}, dynamically decomposing its local structure during training to guide parameter-objective assignment. This shift transforms the Jacobian from a passive monitor into an active control mechanism.

\subsection{Factuality vs. Confidence in LLMs}

Maintaining both factual accuracy and confident output generation is a central concern for large language models~\cite{kadavath2022language}. Prior efforts such as TruthfulQA~\cite{lin2022truthfulqa} and SelfCheckGPT~\cite{manakul2023selfcheckgpt} expose the trade-off between informativeness and entropy. Entropy filtering~\cite{lee2022factuality} and output calibration have been proposed as inference-time remedies, but they do not intervene during model optimization.

Our method addresses this gap by directly incorporating the dual objectives of \texttt{fidelity} (via cross-entropy) and \texttt{certainty} (via entropy minimization) into the training process. Using Jacobian-based attribution, we enable the model to resolve their competition adaptively during learning, rather than relying on fixed heuristics or post-hoc adjustments.




\section{Method}

We propose the \textbf{J6 Strategy}, a Jacobian-guided optimization framework for multi-objective prompt tuning. Our method aims to jointly optimize two often competing goals in large language model (LLM) inference: \textbf{fidelity} (i.e., task accuracy, denoted as \texttt{Heat}) and \textbf{certainty} (i.e., output confidence, denoted as \texttt{Confidence}). Unlike traditional approaches that bind each objective to specific parameters or scalarize multiple losses into a single function, J6 introduces a structured and interpretable mechanism for dynamically allocating optimization responsibility across parameter groups.

We begin by formalizing the problem setup and highlighting the limitations of static role assignments in multi-objective tuning. We then introduce the \textit{Jacobian Interaction Matrix}—a local $2 \times 2$ matrix capturing how each objective gradient aligns with each parameter direction—and decompose it into the \textbf{J6} score vector. This vector encodes role-specific effectiveness and cross-objective alignment.

Finally, we describe two optimization modes built a top J6: a \textit{hard routing} strategy that selects dominant roles via argmax, and a \textit{soft weighting} strategy that adaptively blends gradients based on normalized Jacobian scores. Together, these components enable flexible, conflict-aware updates without increasing model dimensionality.




\subsection{Problem Formulation}

We consider a prompt-tuning setting where the goal is to adapt a frozen Large Language Model (LLM) to specific tasks while simultaneously balancing two optimization objectives. To enable lightweight yet expressive adaptation, we introduce two additive parameter groups:
\begin{itemize}
    \item $h \in \mathbb{R}^{d_h}$: perturbations applied to the hidden layers,
    \item $w \in \mathbb{R}^{d_w}$: perturbations applied to the vocabulary embedding layer.
\end{itemize}

These parameters modulate the model's output logits, which we approximate as:
\[
\text{logits} = (H + h)(W + w)^{T},
\]
where $H$ and $W$ represent the frozen base representations (i.e., the hidden states and vocabulary embeddings of the pretrained LLM), and $(h, w)$ are the tunable low-rank adaptations. Specifically, $W$ corresponds to the pre-trained vocabulary embeddings, and the perturbation $w$ is added to each embedding vector, adjusting how each token interacts with the hidden state.

We define the objectives as:
\begin{align}
\text{ob}_1(h, w) &= \mathrm{CrossEntropy}(\text{logits},\ y) \quad \text{(Heat)}~\text{\cite{hu2025slotsamplespecificlanguagemodel}} \label{eq:heat} \\ 
\text{ob}_2(h, w) &= -\mathrm{Entropy}(\mathrm{softmax}(\text{logits})) \quad \text{(Confidence)}~\text{\cite{agarwal2025unreasonableeffectivenessentropyminimization}} \label{eq:confidence}
\end{align}

The \texttt{Heat} loss is derived from the cross-entropy between the model's logits and the true target, ensuring task alignment. The \texttt{Confidence} loss minimizes the entropy of the softmax probabilities, encouraging confident predictions. In the SLOT framework, this loss can be interpreted as the model's self-prediction of the next token, guiding parameter optimization for task-specific tuning.

However, these objectives often conflict. In tasks requiring reasoning or domain adaptation, improving accuracy may reduce confidence, and vice versa. Our goal is to design a mechanism that \textit{dynamically balances} these objectives by routing gradient influence through the appropriate parameter group, leveraging geometric attribution.

\subsection{Motivation: Dynamic Role Attribution}

Traditional prompt tuning methods statically assign specific roles to hidden perturbations (\textbf{$h$}) and embedding perturbations (\textbf{$w$})—where \textbf{$h$} is responsible for optimizing \texttt{Heat} (cross-entropy) and \textbf{$w$} optimizes \texttt{Confidence} (entropy minimization). However, this rigid assignment often fails to capture the dynamic interactions between these two tasks in practice.

Given the multiplicative structure of the logits in large language models (LLMs):
\[
\text{logits} = (H + h)(W + w)^{T},
\]
updates to \textbf{$h$} and \textbf{$w$} are inherently entangled—each parameter can influence both objectives in complex, unpredictable ways. For example, adjusting \textbf{$h$} for improved accuracy can inadvertently affect the confidence, and modifying \textbf{$w$} to enhance certainty can potentially undermine the accuracy.

To address these limitations, we propose a dynamic role attribution framework that eliminates the need for fixed roles. Instead of predefining tasks for each parameter group, we allow \textbf{$h$} and \textbf{$w$} to dynamically choose their optimization strategy at each step. They can choose to support \texttt{Heat}, \texttt{Confidence}, both, or remain inactive, depending on which task they can most effectively contribute to at that moment.

This dynamic decision-making process is guided by the local gradient landscape. Specifically, we construct a \textbf{Jacobian Interaction Matrix} that quantifies how each parameter influences each objective. From this matrix, we derive the \textbf{J6 score}, a fine-grained signal that enables either competitive or cooperative updates, achieved through discrete strategy routing or soft weighting.

In this way, \textbf{J6} overcomes the limitations of brittle, pre-defined heuristics by providing interpretable, geometry-aware role decisions, ultimately enabling conflict-aware prompt optimization under multi-objective supervision~\cite{yu2020gradient, liu2021conflict}.

\subsection{Jacobian Construction and J6 Score Vector}

To capture how each group of prompt parameters affects each optimization objective, we construct the local Jacobian matrix $J \in \mathbb{R}^{2 \times 2}$ at each step of training. It summarizes the partial derivatives of two objectives with respect to two parameter groups:

\[
J = 
\begin{bmatrix}
J_{11} & J_{12} \\
J_{21} & J_{22}
\end{bmatrix}, \quad \text{where} \quad J_{ij} = \nabla_{\theta_j} \text{ob}_i,\quad \theta_1 = h,\ \theta_2 = w.
\]

We instantiate this Jacobian for our specific objective setting: $\text{ob}_1 = \texttt{Heat}$ (Cross-Entropy), and $\text{ob}_2 = \texttt{Confidence}$ (Negative Entropy). Thus, $J$ becomes:

\[J =
\begin{bmatrix}
\frac{\partial \text{ob}_1}{\partial h} & \frac{\partial \text{ob}_1}{\partial w} \\
\frac{\partial \text{ob}_2}{\partial h} & \frac{\partial \text{ob}_2}{\partial w}
\end{bmatrix}
=
\begin{bmatrix}
\nabla_h \text{Heat} & \nabla_w \text{Heat} \\
\nabla_h \text{Conf} & \nabla_w \text{Conf}
\end{bmatrix}
=
\begin{bmatrix}
J_{11} & J_{12} \\
J_{21} & J_{22}
\end{bmatrix}\]

We now define the \textbf{J6 Score Vector} to decompose the Jacobian into six interpretable components:

\[
    \text{J6} =
    \left[
    \underbrace{\|J_{11}\|^2}_{\text{$h \rightarrow$ Heat}}, \quad
    \underbrace{\|J_{12}\|^2}_{\text{$w \rightarrow$ Heat}}, \quad
    \underbrace{\|J_{21}\|^2}_{\text{$h \rightarrow$ Confidence}}, \quad
    \underbrace{\|J_{22}\|^2}_{\text{$w \rightarrow$ Confidence}}, \quad
    \underbrace{\langle J_{11}, J_{22} \rangle}_{\text{Cross Alignment}}, \quad
    \underbrace{\langle J_{21}, J_{12} \rangle}_{\text{Cross Alignment}}
    \right]
\]

Each component reflects a specific geometric or optimization property of the parameter-objective landscape, as summarized in the table below:

\begin{table}[h]
    \centering
    \caption{Interpretation of each J6 score component. $\|J_{ij}\|^2$ quantifies parameter-objective influence; inner products reflect alignment.}
    \label{tab:j6_component_table}
    \vspace{0.5em}
    \begin{tabular}{|c|l|l|}
    \hline
    \textbf{J6 Component} & \textbf{Gradient Role} & \textbf{Optimization Interpretation} \\
    \hline
    $\|J_{11}\|^2$ & $h$'s contribution to Heat & $h \rightarrow \text{Heat}$ \\
    $\|J_{22}\|^2$ & $w$'s contribution to Confidence & $w \rightarrow \text{Confidence}$ \\
    $\|J_{12}\|^2$ & $w$'s influence on Heat & $w \rightarrow \text{Heat}$ \\\
    $\|J_{21}\|^2$ & $h$'s influence on Confidence & $h \rightarrow \text{Confidence}$ \\\
    $\langle J_{11}, J_{22} \rangle$ & Cross-path interaction & $h \rightarrow \text{Heat}$ \, and $w \rightarrow \text{Confidence}$ \, coupling \\
    $\langle J_{21}, J_{12} \rangle$ & Cross-path interaction & $h \rightarrow \text{Confidence}$ \, and $w \rightarrow \text{Heat}$ \, coupling \\
    \hline
    \end{tabular}
\end{table}


Each component quantifies a distinct aspect of the local interaction geometry, including how each parameter group contributes to its primary objective, how it interferes with the other, and whether the two groups align or conflict in their gradient directions. These interpretations are summarized in Table~\ref{tab:j6_component_table}.

This decomposition enables interpretable and structured decision-making in prompt optimization. Rather than applying static or uniform updates to $h$ and $w$, J6 provides a geometry-aware signal that supports two optimization strategies:

\begin{itemize}
    \item \textbf{Hard Strategy}: discretely selects one of six roles per parameter group at each step (e.g., "optimize for Heat", "remain frozen").
    \item \textbf{Soft Strategy}: continuously reweights objective gradients using J6-aligned scores, enabling smooth interpolation and partial cooperation across objectives.
\end{itemize}

  We formalize this in the following optimization algorithm, which can be invoked at each training or adaptation step for a given prompt instance.
  
  \begin{algorithm}[H]
    \caption{J6 Slot Optimization and Inference for a Given Prompt}
    \label{alg:j6_slot}
    \begin{algorithmic}[1]
      \REQUIRE Base representations $H$, $W$; tunable parameters $h$, $w$; input prompt $x$ and target $y$.
      \STATE Compute logits: $\text{logits} = (H + h)(W + w)^T$
      \STATE Compute losses: $\text{ob}_1$ (Heat), $\text{ob}_2$ (Confidence)
      \STATE Compute gradients: $\nabla_h \text{ob}_1$, $\nabla_h \text{ob}_2$, $\nabla_w \text{ob}_1$, $\nabla_w \text{ob}_2$
      \STATE Construct Jacobian blocks: $J_{11}, J_{12}, J_{21}, J_{22}$
      \STATE Compute J6 score vector: 
      \[
      \text{J6} = 
      \left[
      \|J_{11}\|^2,\;
      \langle J_{11}, J_{22} \rangle,\;
      \|J_{12}\|^2,\;
      \|J_{21}\|^2,\;
      \|J_{22}\|^2,\;
      \langle J_{21}, J_{12} \rangle
      \right]
      \]
      \IF{Hard Routing}
          \STATE Assign update role to $h$ and $w$ based on max-scoring J6 slot
          \STATE Apply gradient update only along selected objective direction(s)
      \ELSIF{Soft Weighting}
          \STATE Normalize J6 components to obtain gradient weights $\alpha_1, \alpha_2, \ldots$
          \STATE Combine gradients: $\nabla_h = \alpha_h^{(1)} \nabla_h \text{ob}_1 + \alpha_h^{(2)} \nabla_h \text{ob}_2$
          \STATE Similarly compute weighted $\nabla_w$
      \ENDIF
      \STATE Update $h$, $w$ via optimizer step
    \end{algorithmic}
  \end{algorithm}
  
  \subsection{J$^+$: Fine-Grained Attribution via Expanded Jacobian}

  The J6 vector provides a compact representation of the key roles of $h$ and $w$ through six gradient metrics. While useful, it offers only a coarse partition of the parameter-objective interactions and misses out on capturing more complex behaviors, such as when $h$ supports both Heat and Confidence simultaneously, or when $w$ aids one objective while conflicting with the other.
  
  \paragraph{From J6 to J$^+$.}
  Although J6 captures the key gradient interactions between perturbation parameters $h$ and $w$, it serves as a low-dimensional approximation of the full gradient interaction space encoded by J$^+$. While J6 retains essential components like primary gradient magnitudes, interference terms, and alignment scores, it cannot fully represent the complexity of $h$ and $w$'s interactions, particularly when they simultaneously affect multiple objectives in non-linear ways.
  
  To address these limitations, we propose \textbf{J$^+$}, an extended Jacobian-based scoring system that decomposes the full $2 \times 2$ gradient Jacobian matrix $J$ into 15 interpretable components. These components enumerate all possible gradient magnitudes, alignments, and cross-objective contributions, thus providing a more granular and expressive optimization framework.

  The \textbf{J$^+$ space} consists of $4 \times 4 - 1 = 15$ interpretable terms, where the "4" represents the four directional roles for each parameter group: $h_1$, $h_2$, $h_1 + h_2$, and None for $h$; and similarly for $w$. The subtraction of 1 removes the degenerate case (None--None), which is uninformative in optimization.

  These 15 interactions represent meaningful directional relationships between $h$ and $w$, enabling both discrete and soft-weighted optimization strategies based on the gradient geometry.
  
\begin{table}[h]
    \centering
    \caption{J$^+$ Gradient Interaction Matrix: Each cell represents a geometric interaction between $h$ and $w$ directions toward specific objectives.}
    \vspace{0.5em}
    \renewcommand{\arraystretch}{1.4}
    \begin{tabular}{|c|c|c|c|c|}
    \hline
    \diagbox[width=8em]{$h$}{$w$} & $w_1$ (Heat) & $w_2$ (Conf) & $w_1 + w_2$ & None \\
    \hline
    $h_1$ (Heat)
    & $\langle J_{11}, J_{12} \rangle$
    & $\langle J_{11}, J_{22} \rangle$
    & $\langle J_{11}, J_{12} + J_{22} \rangle$
    & $\|J_{11}\|^2$ \\
    \hline
    $h_2$ (Conf)
    & $\langle J_{21}, J_{12} \rangle$
    & $\langle J_{21}, J_{22} \rangle$
    & $\langle J_{21}, J_{12} + J_{22} \rangle$
    & $\|J_{21}\|^2$ \\
    \hline
    $h_1 + h_2$
    & $\langle J_{11}+J_{21}, J_{12} \rangle$
    & $\langle J_{11}+J_{21}, J_{22} \rangle$
    & $\langle J_{11}+J_{21}, J_{12}+J_{22} \rangle$
    & $\|J_{11}+J_{21}\|^2$ \\
    \hline
    None
    & $\|J_{12}\|^2$
    & $\|J_{22}\|^2$
    & $\|J_{12}+J_{22}\|^2$
    & x (invalid) \\
    \hline
    \end{tabular}
    \label{tab:jplus}
\end{table}
  
  Therefore, we can define the J$^+$ vector as:
  \[
    \text{J}^+ = \left[
        \begin{array}{lllll}
        \|J_{11}\|^2, 
        & \|J_{12}\|^2, 
        & \|J_{21}\|^2, 
        & \|J_{22}\|^2, 
        & \langle J_{11}, J_{22} \rangle, \\
        \langle J_{21}, J_{12} \rangle, 
        & \langle J_{11}, J_{21} \rangle, 
        & \langle J_{12}, J_{22} \rangle, 
        & \langle J_{11}+J_{21},\, J_{12}+J_{22} \rangle,
        & \langle J_{11},\, J_{12}+J_{22} \rangle, \\
        \langle J_{21},\, J_{12}+J_{22} \rangle,
        & \langle J_{11}+J_{21},\, J_{12} \rangle,
        & \langle J_{11}+J_{21},\, J_{22} \rangle,
        & \|J_{11}+J_{21}\|^2,
        & \|J_{12}+J_{22}\|^2
        \end{array}
        \right]
  \]
  
  \noindent
  Compared to J6, the J$^+$ vector introduces additional gradient interactions and aggregated behaviors:
  
  \begin{itemize}
    \item $\langle J_{11}, J_{21} \rangle$ — Captures $h$'s internal consistency across \texttt{Heat} and \texttt{Confidence}.
    \item $\langle J_{12}, J_{22} \rangle$ — Captures $w$'s internal consistency across \texttt{Heat} and \texttt{Confidence}.
    
    \item $\langle J_{11}+J_{21}, J_{12}+J_{22} \rangle$ — Measures total coordination between $h$ and $w$ over both objectives.
    \item $\langle J_{11}, J_{12}+J_{22} \rangle$ — Measures how $h$ (on \texttt{Heat}) aligns with overall $w$.
    \item $\langle J_{11}+J_{21}, J_{22} \rangle$ — Measures how overall $h$ aligns with $w$ (on \texttt{Confidence}).
    
    \item $\|J_{11}+J_{21}\|^2$, $\|J_{12}+J_{22}\|^2$ — Aggregate total influence strengths of $h$ and $w$, respectively.

  \end{itemize}

This decomposition provides finer-grained attribution signals, allowing for nuanced role assignments and strategic adaptation in multi-objective tuning. J$^+$ serves as a \textit{superset of J6}, forming the foundation for our soft optimization controller.

    That we can conclude:
    \[\text{J}^+ =
    \left[
    \begin{array}{lllll}
    \underbrace{\|J_{11}\|^2}_{\text{h} \rightarrow \text{Heat}}, &
    \underbrace{\|J_{12}\|^2}_{\text{w} \rightarrow \text{Heat}}, &
    \underbrace{\|J_{21}\|^2}_{\text{h} \rightarrow \text{Conf}}, &
    \underbrace{\|J_{22}\|^2}_{\text{w} \rightarrow \text{Conf}}, &
    \underbrace{\langle J_{11}, J_{22} \rangle}_{\text{Cross Align (h-Heat, w-Conf)}},
    \\
    \underbrace{\langle J_{21}, J_{12} \rangle}_{\text{Cross Align (h-Conf, w-Heat)}}, &
    \underbrace{\langle J_{11}, J_{21} \rangle}_{\text{h → Both}}, &
    \underbrace{\langle J_{12}, J_{22} \rangle}_{\text{w → Both}}, &
    \underbrace{\langle J_{11}+J_{21},\; J_{12}+J_{22} \rangle}_{\text{Joint Align (h + w)}}, &
    \underbrace{\langle J_{11},\; J_{12}+J_{22} \rangle}_{\text{h dominates Heat}},
    \\
    \underbrace{\langle J_{21},\; J_{12}+J_{22} \rangle}_{\text{h dominates Conf}}, &
    \underbrace{\langle J_{11}+J_{21},\; J_{12} \rangle}_{\text{w dominates Heat}}, &
    \underbrace{\langle J_{11}+J_{21},\; J_{22} \rangle}_{\text{w dominates Conf}}, &
    \underbrace{\|J_{11} + J_{21}\|^2}_{\text{h Total Strength}}, &
    \underbrace{\|J_{12} + J_{22}\|^2}_{\text{w Total Strength}}
    \end{array}
    \right]\]
    \noindent
    Given the full Jacobian alignment matrix $J^+ \in \mathbb{R}^{15}$, we define a decision mechanism that selects the dominant optimization pathway by identifying the most aligned gradient channel. Specifically, we compute:
    \[
    \texttt{j\_index} = \arg\max_{j} J^+_j
    \]
    \noindent Based on this matrix, we compile a detailed attribution and decision table to guide both hard and soft strategies under different optimization intents:

    \begin{table}[H]
        \centering
        \renewcommand{\arraystretch}{1.3}
        \caption{\textbf{J$^+$} Attribution and Action Table: Strategy Selection for Hard and Soft Updates}
        \label{tab:jplus_action}
        \begin{tabular}{|c|l|p{4.5cm}|p{4.8cm}|}
        \hline
        \textbf{J Index} & \textbf{Optimization Action} & \textbf{J$^+$ Component} & \textbf{Optimization Suggestion} \\
        \hline
        1  & $h \rightarrow$ Heat                     & $\|J_{11}\|^2$                                          & Optimize $h_1$ only \\
        2  & $w \rightarrow$ Heat                     & $\|J_{12}\|^2$                                          & Optimize $w_1$ only \\
        3  & $h \rightarrow$ Conf                     & $\|J_{21}\|^2$                                          & Optimize $h_2$ only \\
        4  & $w \rightarrow$ Conf                     & $\|J_{22}\|^2$                                          & Optimize $w_2$ only \\
        5  & Cross Align ($h$–Heat, $w$–Conf)         & $\langle J_{11}, J_{22} \rangle$                        & Jointly optimize $h_1$, $w_2$ \\
        6  & Cross Align ($h$–Conf, $w$–Heat)         & $\langle J_{21}, J_{12} \rangle$                        & Jointly optimize $h_2$, $w_1$ \\
        7  & $h \rightarrow$ Both                     & $\langle J_{11}, J_{21} \rangle$                        & Optimize both $h_1 + h_2$ \\
        8  & $w \rightarrow$ Both                     & $\langle J_{12}, J_{22} \rangle$                        & Optimize both $w_1 + w_2$ \\
        9  & Joint Align ($h + w$)                    & $\langle J_{11}+J_{21}, J_{12}+J_{22} \rangle$          & Jointly optimize $h_1 + h_2$, $w_1 + w_2$ \\
        10 & $h$ dominates Heat                       & $\langle J_{11}, J_{12}+J_{22} \rangle$                 & Prioritize $h_1$, auxiliary $w$ \\
        11 & $h$ dominates Conf                       & $\langle J_{21}, J_{12}+J_{22} \rangle$                 & Prioritize $h_2$, auxiliary $w$ \\
        12 & $w$ dominates Heat                       & $\langle J_{11}+J_{21}, J_{12} \rangle$                 & Prioritize $w_1$, auxiliary $h$ \\
        13 & $w$ dominates Conf                       & $\langle J_{11}+J_{21}, J_{22} \rangle$                 & Prioritize $w_2$, auxiliary $h$ \\
        14 & $h$ Total Strength                       & $\|J_{11}+J_{21}\|^2$                                   & Joint optimization over $h_1 + h_2$ \\
        15 & $w$ Total Strength                       & $\|J_{12}+J_{22}\|^2$                                   & Joint optimization over $w_1 + w_2$ \\
        \hline
        \end{tabular}
    \end{table}
    The J$^+$ vector expands upon J6 by offering a finer decomposition of parameter-objective interactions, with 15 terms that cover all gradient alignments and cross-objective contributions. J6, as an orthogonal projection basis of J$^+$, selectively retains the most interpretable components—direct magnitudes, interference terms, and cross-objective alignments—forming a compact yet expressive representation of the local gradient-field dynamics.

    This allows J6 to provide a concise control signal for optimization. By applying a soft-weighted combination of J6 components, we can emulate J$^+$'s directional guidance without the full dimensional cost, enabling both discrete and differentiable strategies for conflict-aware parameter updates.
    
\subsection{Soft (Weighted) Strategy}
    
    To enhance optimization flexibility and interpretability, we introduce a \textbf{soft (weighted) update strategy} based on the geometry of the Jacobian matrix. This strategy is grounded in the relationship between the \textbf{J6 vector} and the full \textbf{J$^+$ interaction matrix} (Table~\ref{tab:jplus}).
    
    \paragraph{From J$^+$ to J6.}
    The J$^+$ matrix represents all meaningful pairwise interactions between perturbation directions $h$, $w$ and objectives (Heat, Confidence). It spans a 4×4 interaction space including alignment, interference, and composite behaviors. However, many of these components are interdependent or redundant in optimization.
    
    In contrast, the J6 vector is constructed as a low-dimensional projection of J$^+$—specifically chosen for its semantic interpretability and optimization utility. It extracts six components: primary gradients ($\|J_{11}\|^2$, $\|J_{22}\|^2$), cross-objective interferences ($\|J_{12}\|^2$, $\|J_{21}\|^2$), and synergy terms ($\langle J_{11}, J_{22} \rangle$, $\langle J_{21}, J_{12} \rangle$).
    
    \vspace{0.5em}
    This geometric foundation enables us to construct a differentiable soft routing mechanism over the J6 vector, allowing parameter updates to \textbf{adaptively emphasize useful directions} while remaining fully trainable. Our soft strategy thus acts as a low-rank controller that approximates the full behavior space of J$^+$ through interpretable, learnable signal composition.

    \paragraph{Step 1: Temperature-scaled weighting.}
    We begin by applying a softmax with temperature $\tau$ to the raw J6 scores:
    \[
    \tilde{\alpha}_i = \frac{\exp(\text{J6}[i]/\tau)}{\sum_j \exp(\text{J6}[j]/\tau)},
    \]
    where $\tau$ controls the sharpness of selection. A small $\tau$ leads to near-argmax behavior (hard attention), while a large $\tau$ yields more uniform weighting.
    
    \paragraph{Step 2: Competitive contrast enhancement.}
    To further amplify high-scoring components while preserving smoothness, we apply a \textit{competitive reward operator}, squaring each weight and re-normalizing:
    \[
    \alpha_i = \frac{\tilde{\alpha}_i^\gamma}{\sum_j \tilde{\alpha}_j^\gamma}, \quad \text{where } \gamma > 1.
    \]
    This enhances contrast (akin to energy-based methods) without excluding weaker signals. For example, $(0.5, 0.3, 0.1, 0.1) \rightarrow (0.25, 0.09, 0.01, 0.01)$ after squaring.
    
    \paragraph{Step 3: Weighted updates.}
    Finally, the parameters are updated via weighted gradients over multiple objectives:
    \[
    \Delta h = -\eta_h \left(
    \alpha_0 \nabla_h \text{ob}_1 +
    \alpha_3 \nabla_h \text{ob}_2
    \right), \qquad
    \Delta w = -\eta_w \left(
    \alpha_2 \nabla_w \text{ob}_1 +
    \alpha_4 \nabla_w \text{ob}_2
    \right),
    \]
    where $\text{ob}_1$, $\text{ob}_2$ represent Heat and Confidence objectives respectively. Additional components $\alpha_1$ and $\alpha_5$ can scale auxiliary updates, such as alignment penalties or entropy regularization.
    
    \vspace{0.5em}
    This soft strategy offers a continuous and interpretable optimization mechanism. It balances objectives dynamically using geometric signals, and encourages diverse directional exploration while maintaining differentiability.
    
    \begin{algorithm}[H]
    \caption{J6 Weighted Update: Soft Strategy}
    \label{alg:j6_soft}
    \begin{algorithmic}[1]
    \REQUIRE J6 vector $\text{J6} \in \mathbb{R}^6$, temperature $\tau$, exponent $\gamma$
    \STATE $\tilde{\alpha}_i \gets \exp(\text{J6}[i]/\tau)$ for $i = 0,\dots,5$
    \STATE Normalize: $\tilde{\alpha}_i \gets \tilde{\alpha}_i / \sum_j \tilde{\alpha}_j$
    \STATE Apply reward operator: $\alpha_i \gets \tilde{\alpha}_i^\gamma$
    \STATE Renormalize: $\alpha_i \gets \alpha_i / \sum_j \alpha_j$
    \STATE Compute update for $h$: $\Delta h \gets -\eta_h (\alpha_0 \nabla_h \text{ob}_1 + \alpha_3 \nabla_h \text{ob}_2)$
    \STATE Compute update for $w$: $\Delta w \gets -\eta_w (\alpha_2 \nabla_w \text{ob}_1 + \alpha_4 \nabla_w \text{ob}_2)$
    \RETURN $\Delta h$, $\Delta w$
    \end{algorithmic}
    \end{algorithm}

\section{Conclusion}

In this paper, we introduced J6, a structured Jacobian-based approach for multi-objective prompt optimization in large language models (LLMs). We identified the fundamental challenge of balancing multiple, often conflicting, objectives—specifically fidelity (accuracy) and certainty (confidence)—in the context of prompt tuning. Traditional methods fall short in modeling the complex interactions between parameters and objectives, typically relying on scalar aggregation techniques that ignore the underlying gradient geometry.

Our proposed solution, J6, decomposes the gradient interaction matrix into six interpretable components, providing a clear framework for dynamic role attribution. By using hard and soft strategies, J6 offers both discrete role assignments and continuous gradient blending, allowing the model to adaptively respond to local conflicts and synergies. This enables a more flexible, interpretable, and effective optimization process compared to existing methods.

We extended this approach to J+, a richer Jacobian space that incorporates 15 alignment terms, offering even finer granularity. 

Our work presents a novel direction for integrating Jacobian-based reasoning into prompt optimization, offering both theoretical insight and practical performance gains. The J6 framework not only provides a more effective means of balancing accuracy and confidence but also opens the door for further exploration of structured gradient reasoning in neural network optimization.

\textbf{Future work} can explore extending J6 to even larger models, incorporating more complex multi-objective settings, and further refining the soft weighting strategies for finer control. Additionally, exploring the application of J6 in other areas such as reinforcement learning or multi-task learning could provide valuable insights into its broader applicability.

\bibliographystyle{plainnat}
\bibliography{ref}

\end{document}